\documentclass[twoside,11pt]{article}

\usepackage{jmlr2e}

\usepackage{hyperref}

\usepackage{color}

\usepackage{url}
\usepackage{hyperref}
\usepackage{color}

\usepackage{wrapfig}
\usepackage{epsfig}
\usepackage{graphicx}

\usepackage{amssymb}

\usepackage{enumerate}

\usepackage{natbib}

\ShortHeadings{F$^*$: An interpretable transformation of the
  F-measure}{Hand, Christen and Kirielle}

\firstpageno{1}

\begin{document}


\title{F*: An Interpretable Transformation of the F-measure}

\author{\name David J. Hand \email d.j.hand@imperial.ac.uk \\
       \addr Department of Mathematics \\
       Imperial College London, London, UK
       \AND
       \name Peter Christen \email peter.christen@anu.edu.au \\
       \name Nishadi Kirielle \email nishadi.kirielle@anu.edu.au \\
       \addr School of Computing \\
       The Australian National University, Canberra, Australia}

\editor{}

\maketitle

\begin{abstract}
The F-measure, also known as the F1-score, is widely used to assess
the performance of classification algorithms. However, some
researchers find it lacking in intuitive interpretation, questioning
the appropriateness of combining two aspects of performance as
conceptually distinct as precision and recall, and also questioning
whether the harmonic mean is the best way to combine them. To ease
this concern, we describe a simple transformation of the F-measure,
which we call $F^*$ (F-star), which has an immediate practical
interpretation.
\end{abstract}



\section{Introduction}

Many different measures have been used to evaluate the performance
of classification algorithms~\citep[see, for
example,][]{Chi20,Dem06,Fer09,Han12,Pow11,Sok09}. 
Such evaluation is central to choosing between algorithms -- to
decide which is the best to use in practice, to decide if a method
is ``good enough", to optimise parameters (equivalent to choosing
between methods), and for other reasons. The data on which such
assessments are based is normally a test set that is independent
of the training data, consisting of a score and an associated true
class label for each object. Here we consider the two-class case,
with labels 0 and 1. Objects are assigned to class 1 if their score
exceeds some threshold $t$, and to class 0 otherwise. This reduces
the data for the evaluation measure to a two-by-two table, the
confusion matrix, with counts as shown in
Table~\ref{tab-confusion-matrix}.

\renewcommand{\arraystretch}{1.2}

\begin{table}[th!]
  \centering
  \begin{tabular}{|c|c|c|c|} \cline{3-4}
  \multicolumn{2}{c}{~} & \multicolumn{2}{|c|}{\emph{True class}}
    \\ \cline{3-4}
  \multicolumn{2}{c|}{~} & 0 & 1 \\ \hline
  ~\emph{Predicted}~ & 0 & ~$TN$ (true negatives)~ & ~$FN$ (false
    negatives)~ \\ \cline{2-4}
  ~\emph{class}~ & 1 & ~$FP$ (false positives)~ & ~$TP$ (true
    positives)~ \\ \hline
  \end{tabular}
  \caption{Notation for confusion matrix.}
  \label{tab-confusion-matrix}
\end{table}

In general, such a table has four degrees of freedom. Normally,
however, the total number of test set cases, $n=TN+FN+FP+TP$, will
be known, as will the relative proportions belonging to each of the
two classes. These are sometimes called the priors, or the prevalence
in medical applications. This reduces the problem to just two
degrees of freedom, which must be combined in some way in order to
yield a numerical measure on a univariate continuum which can be
used to compare classifiers. The choice of the two degrees of
freedom and the way of combining them can be made in various ways.
In particular, the columns and rows of the table yield proportions
which can then be combined (using the known relative class sizes).
These proportions go under various names, including, recall or
sensitivity, $TP/(TP+FN)$; precision or positive predictive value,
$TP/(TP+FP)$; specificity, $TN/(TN+FP)$; and negative predictive
value, $TN/(TN+FN)$.

These simple proportions can be combined to yield familiar
performance measures, including the misclassification rate, the
kappa statistic, the Youden index, the Matthews coefficient, and
the F-measure or F1-score~\citep{Chi20,Han12}.

Another class of measures acknowledges that the value of the
classification threshold $t$ which is to be used in practice may
not be known at the time that the algorithm has to be evaluated
and when a choice between algorithms has to be made, so that they
average over a distribution of possible values of $t$. Such
measures include the Area Under the Receiver Operating
Characteristic Curve (AUC)~\citep{Dav06} and the
H-measure~\citep{Han09,Han14}.

We should remark that the various names are not always used
consistently and also that particular measures go under different
names, this being a consequence of the widespread applications of
the ideas, which arise in many different application domains. An
example is the equivalence of recall and sensitivity discussed
above.

Many of the performance measures have straightforward intuitive
interpretations. For example:
\begin{itemize}
\item the misclassification rate is simply the proportion of
  objects in the test set which are incorrectly classified;
\item the kappa statistic is the chance-adjusted proportion
  correctly classified;
\item the AUC is the probability that a randomly chosen class 0
  object will have a score lower than a randomly chosen class 1
  object; and
\item the H-measure is the fraction by which the classifier
  reduces the expected minimum misclassification loss, compared
  with that of a random classifier.
\end{itemize}

The F-measure is particularly widely used in computational
disciplines. It was originally developed in the context of
information retrieval to evaluate the ranking of documents retrieved
based on a query~\citep{Van79}. In recent times the F-measure has
gained increasing interest in the context of classification,
especially to evaluate imbalanced classification problems, in
various domains including machine learning, computer vision, data
analytics, and natural language processing. It has a simple
interpretation as the harmonic mean of the two confusion matrix
degrees of freedom precision, $P=TP/(TP+FP)$, and recall,
$R=TP/(TP+FN)$:
\begin{equation}
  F = \frac{2}{\frac{1}{P} + \frac{1}{R}} = \frac{2 P R }{P + R}.
  \label{eqn:f}
\end{equation}

Since precision and recall tap different, and in a sense
complementary, aspects of classification performance, it seems
reasonable to combine them into a single measure. But averaging them
may not be so palatable. One can think of precision as an empirical
estimate of the conditional probability of a correct classification
given predicted class 1 ($Prob(True=1|Pred=1)$), and recall as being
an empirical estimate of the conditional probability of a correct
classification given true class 1 ($Prob(Pred=1|True=1)$). An
average of these has no interpretation as a probability.

Moreover, despite the seminal work of
\citet{Van79}, some researchers are uneasy about the use of the
harmonic mean~\citep{Han18}, preferring other forms of average
(e.g. an arithmetic or geometric mean). For example, the harmonic
mean of two values has the property that it lies closer to the
smaller of the values than the larger. In particular, if one of
recall or precision is zero, then the harmonic mean (and therefore
$F$) is zero, ignoring the value of the other. The desire for an
interpretable perspective on $F$ has been discussed, for example,
on~\citet{Sta13}. 

In an attempt to tackle this unease, in what follows we present a
transformed version of the F-measure which has a straightforward
intuitive interpretation.


\section{The F-measure and F*}

Plugging the counts from Table~\ref{tab-confusion-matrix} into the
definition of $F$, we obtain
$$
  F = \frac{2}{\frac{TP+FP}{TP} + \frac{TP+FN}{TP}} =
      \frac{2 TP}{FN+FP+2TP},
$$
from which
$$
  \frac{TP}{FN+FP} = \frac{1}{2} \frac{F}{1-F}.
$$

So if we define $F'$ as $F' = F / 2(1-F)$, we have that $F'$ is
\emph{the number of class 1 objects correctly classified for each
object that is misclassified}.


This is a straightforward and attractive interpretation of a
transformation of the F-measure, and some researchers might prefer
to use it. However, $F'$ has the property that it is a ratio and
not simply a proportion, so it is not constrained to lie between 0
and 1 -- as are most other performance measures. 

We can overcome this by a further transformation, yielding
\begin{equation}
  \frac{TP}{FN+FP+TP} = \frac{F}{2-F}.
  \label{eqn:fstar}
\end{equation}
Now, defining $F^*$ (F-star) as $F^* = F/(2-F)$\footnote{In terms of
precision and recall, $F^* = P R / (P+R-P R)$.}, we have that:
\begin{center} 
  \begin{tabular}{p{0.9\textwidth}}
  \textbf{\emph{F$^*$ is the proportion of the relevant classifications
  which are correct, where a relevant classification is one which is
  either really class 1 or classified as class 1}.}
  \end{tabular}
\end{center}
Under some circumstances, researchers might find alternative ways of
looking at $F^*$ useful. In particular:
\begin{itemize}
\item $F^*$ is \emph{the number of correctly classified class 1
  objects expressed as a fraction of the number of objects which are
  either misclassified or are correctly classified class 1 objects};
  or,
\item $F^*$ is \emph{the number of correctly classified class 1
  objects expressed as a proportion of the number of objects which
  are either class 1, classified as class 1, or both}; or, yet a
  fourth alternative,
\item $F^*$ is \emph{the number of correctly classified class 1
  objects expressed as a fraction of the number of objects which
  are not correctly classified class 0 objects}.
\end{itemize}

$F^*$ can be alternatively written as $F^*=TP/(n-TN)$, which can be
directly calculated from the confusion matrix.
%
Researchers may recognise this as the Jaccard coefficient, widely
used in areas where true negatives may not be relevant, such as
numerical taxonomy and fraud analytics~\citep{Jac08,Dun82,Bae15}.  

\begin{wrapfigure}[15]{r}{0.35\textwidth}
  \centering
  \includegraphics[width=0.28\textwidth]
    {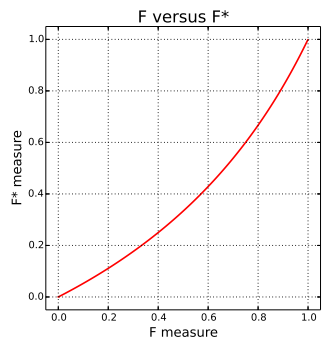}
  \caption{The transformation from $F$ to $F^*$.
    \label{fig:relationship}}
\end{wrapfigure}

To illustrate, if class 1 objects are documents in information
retrieval, then $F^*$ is the number of relevant documents retrieved
expressed as a proportion of all documents except non-retrieved
irrelevant documents. Or, if class 1 objects are COVID-19 infections,
then $F^*$ is the number of infected people who test positive divided
by the number who either test positive or are infected or both.

\begin{figure}[t]
  \centering
  \includegraphics[width=0.28\textwidth]
    {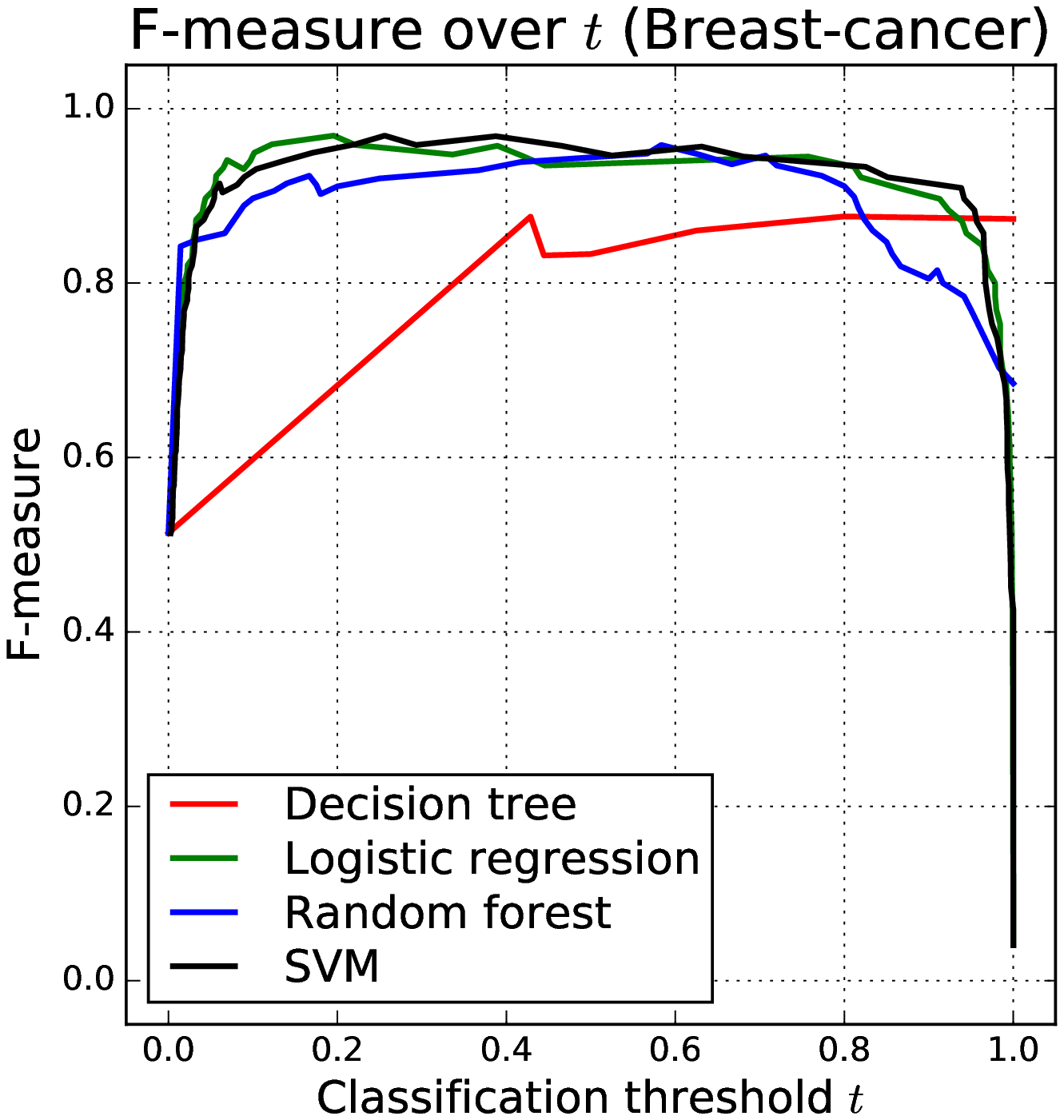} ~~~~~
  \includegraphics[width=0.28\textwidth]
    {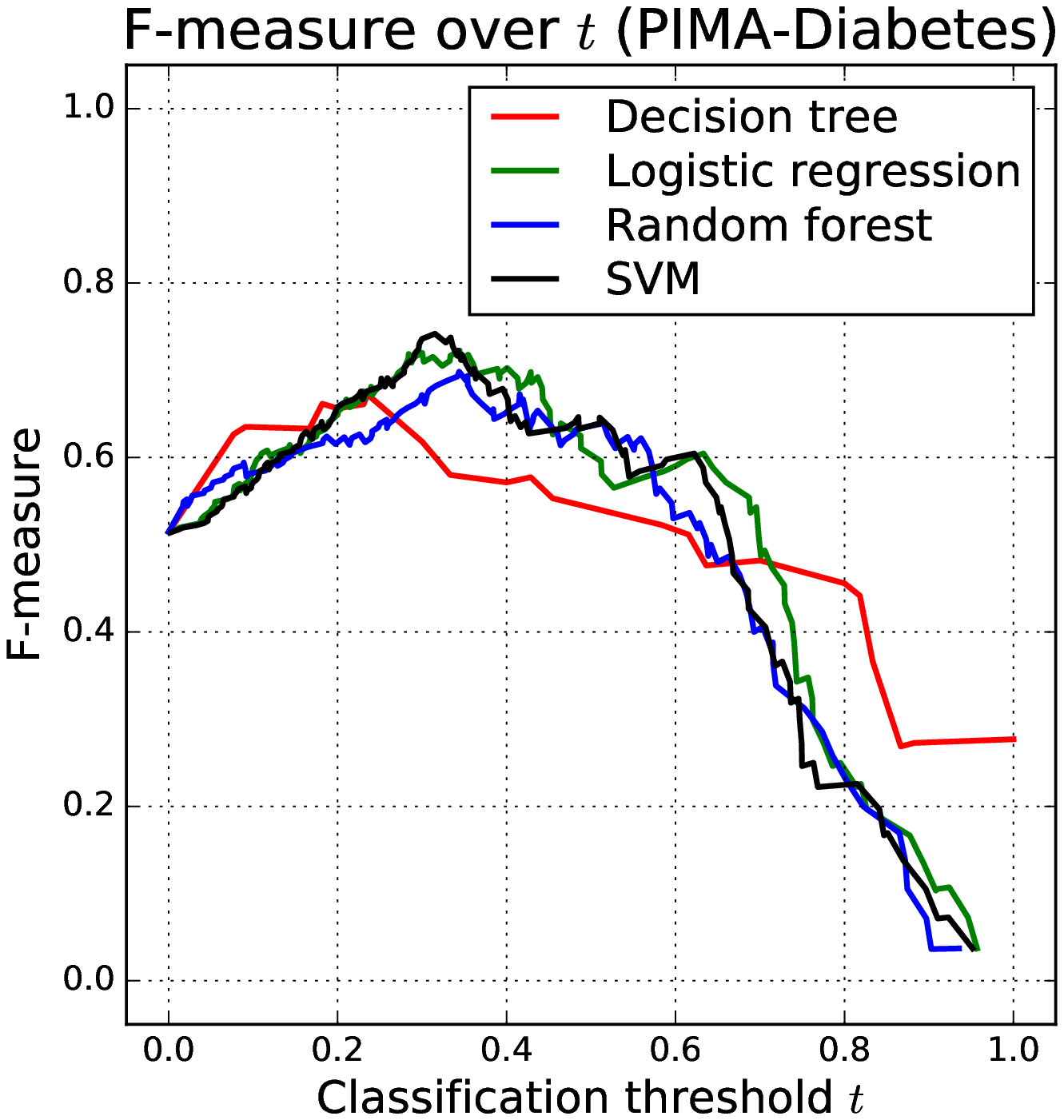} ~~~~~
  \includegraphics[width=0.28\textwidth]
    {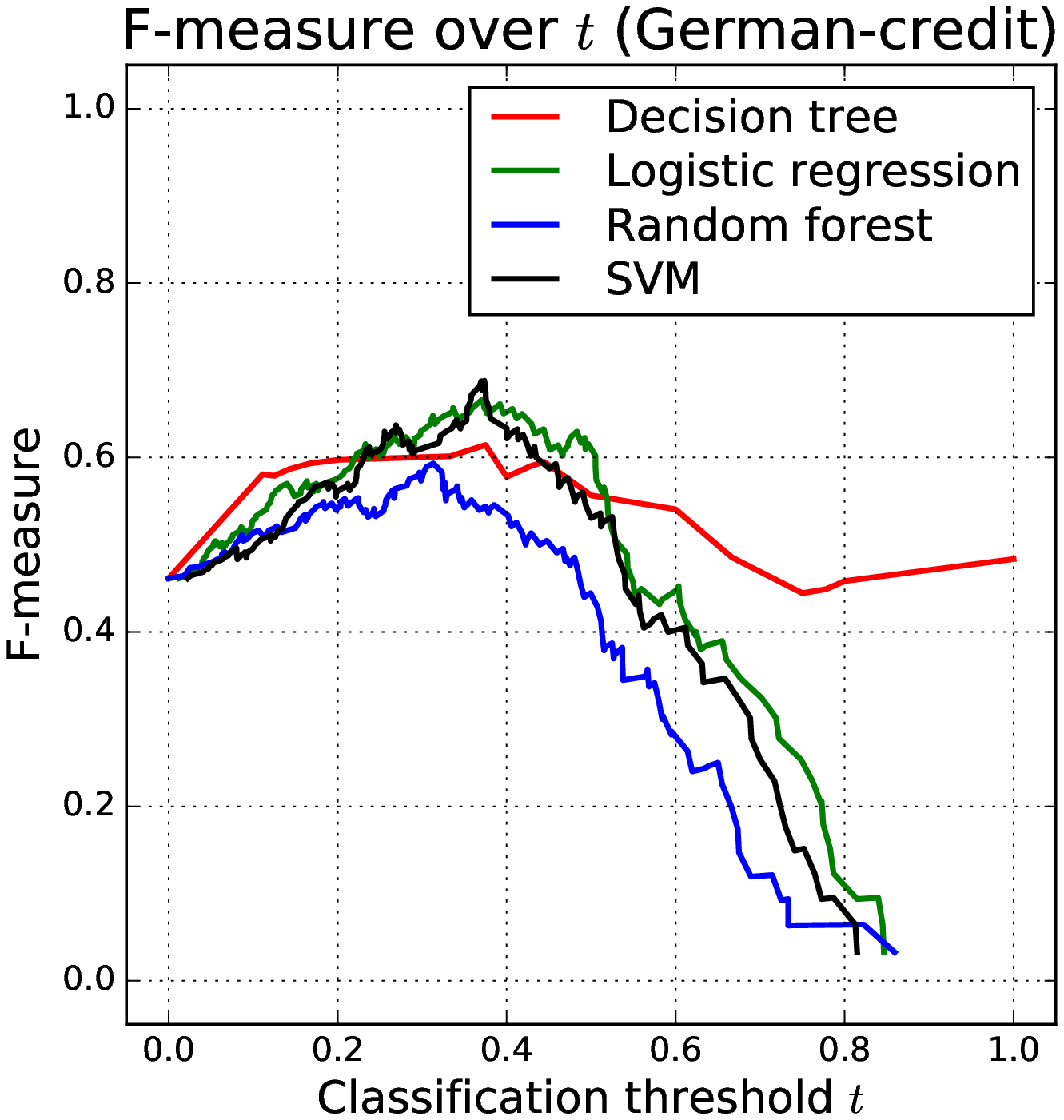}
  \includegraphics[width=0.28\textwidth]
    {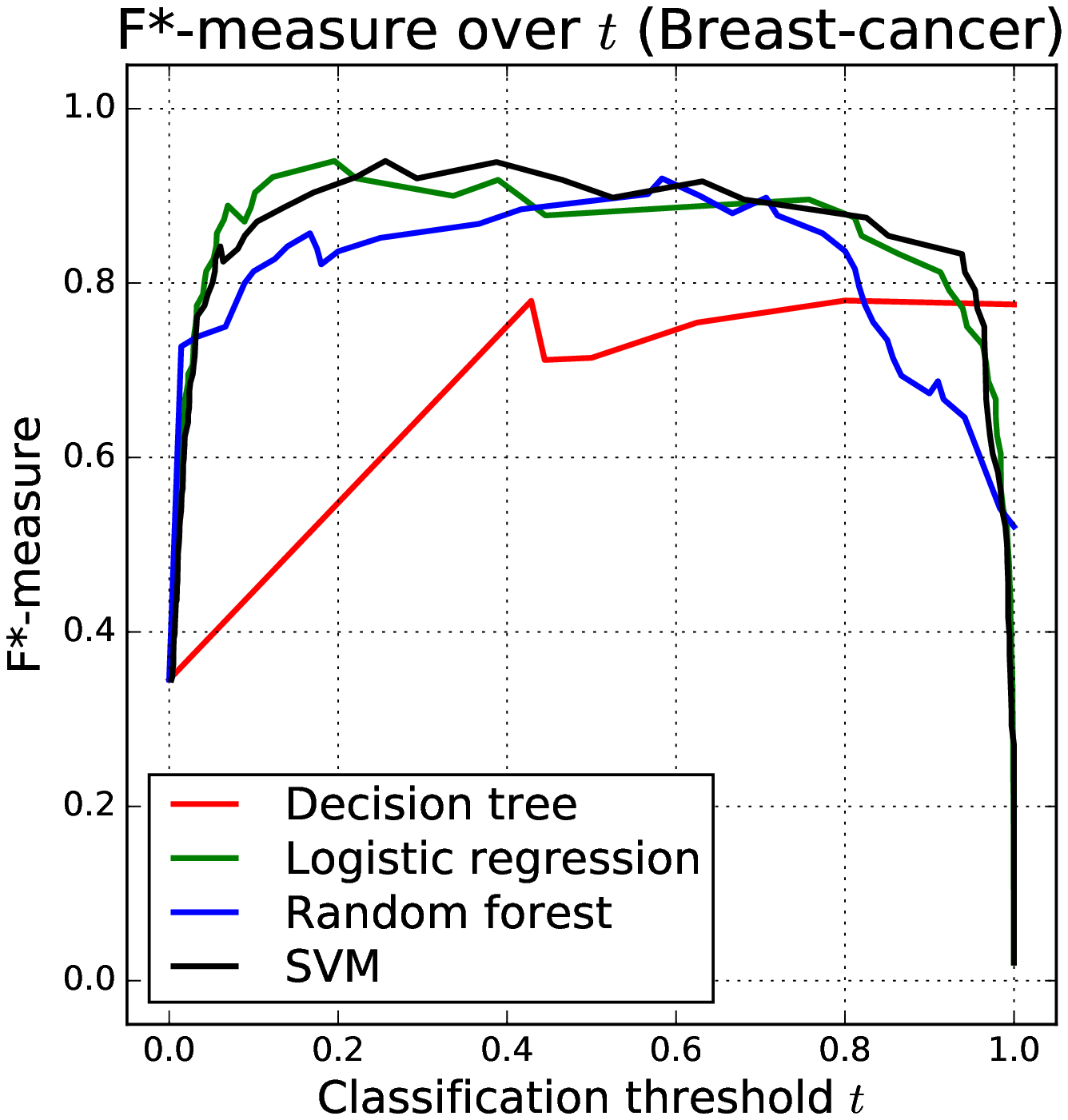} ~~~~~
  \includegraphics[width=0.28\textwidth]
    {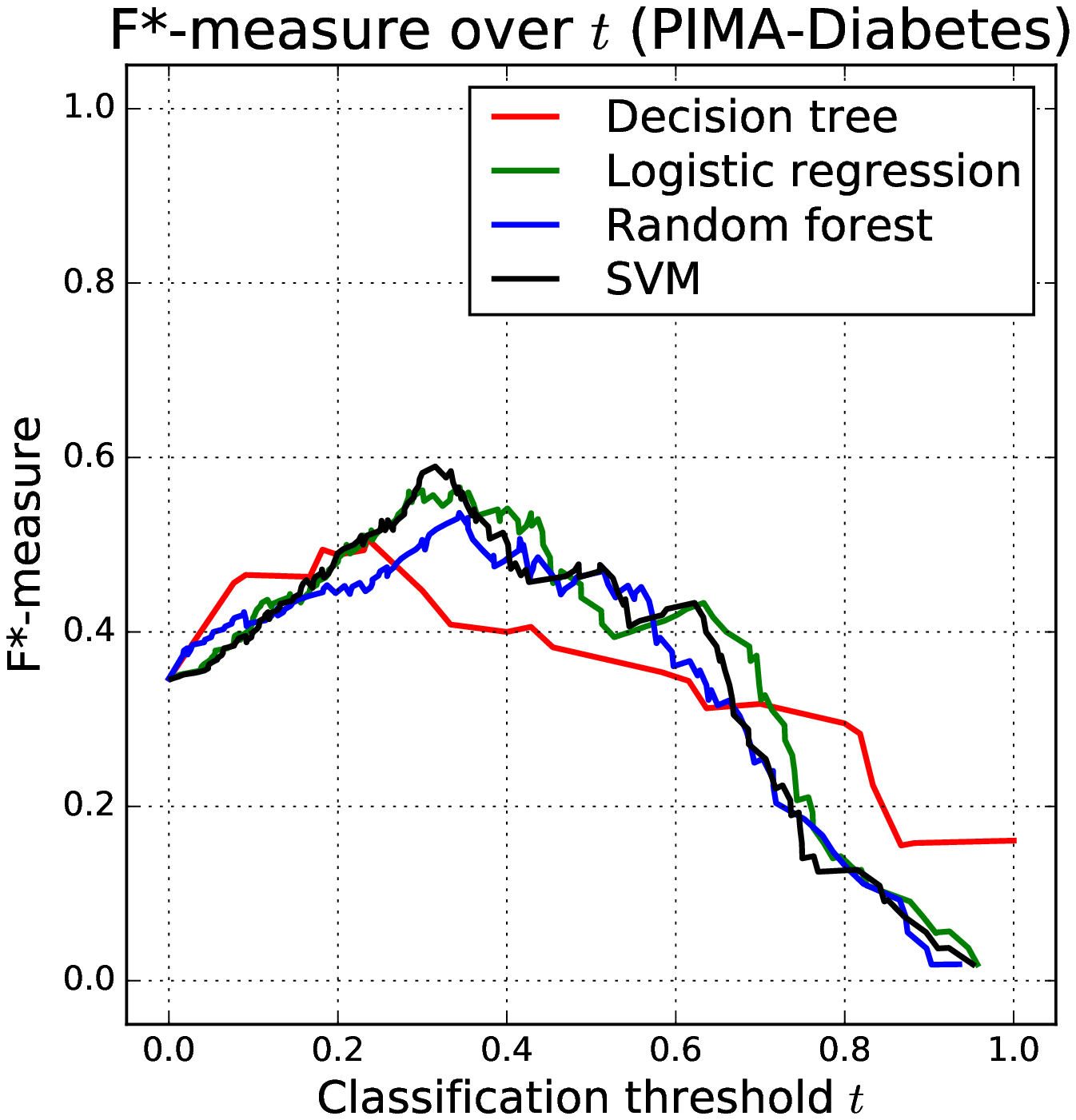} ~~~~~
  \includegraphics[width=0.28\textwidth]
    {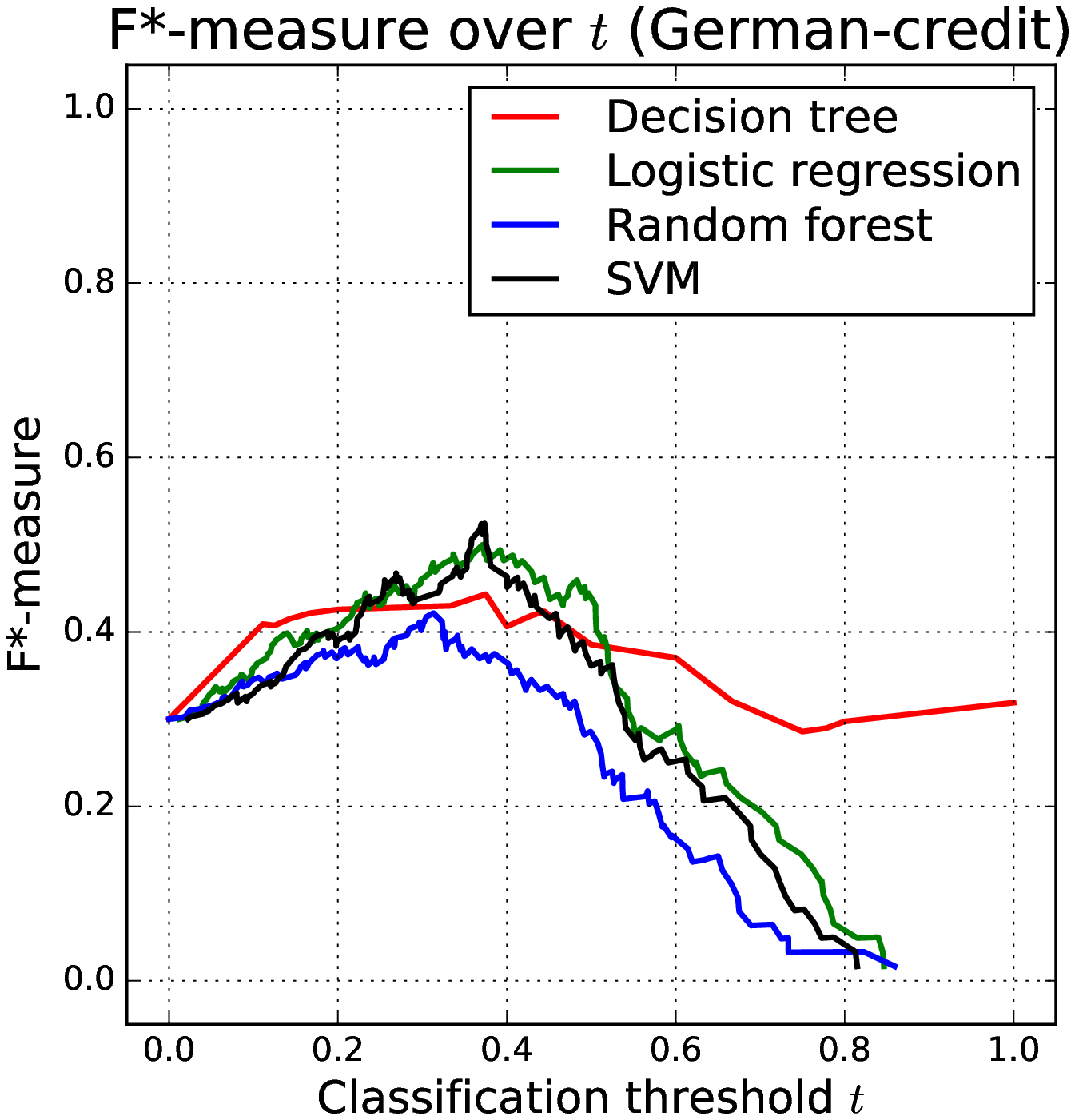}    
  \caption{Experimental results showing the $F$ (top) and $F^*$
    (bottom) measure on three public data sets using four
    classification techniques.
    \label{fig:data-plots}}
\end{figure}

The relationship between $F^*$ and $F$ is shown in
Figure~\ref{fig:relationship}. The approximate linearity of this
curve shows that $F^*$ values will be close to $F$ values. More
importantly, however, is the fact that $F^*$ is a monotonic
transformation of $F$. This means that any conclusions reached by
seeing which $F^*$ values are larger will be identical to those
reached by seeing which $F$ values are larger. In particular, choices
between algorithms will be the same. This is illustrated in
Figure~\ref{fig:data-plots}, which shows experimental results for
three public data sets from the UCI Machine Learning
Repository~\citep{Lic13} for four classifiers as implemented using
Sklearn~\citep{Ped11} with default parameter settings and the
classification threshold $t$ varying between 0 and 1. Although the
curve shapes differ slightly between $F^*$ and $F$ (because of the
monotonic $F$ to $F^*$ transformation of the vertical axis), the
threshold values at which they cross are the same.

\citet{Van79} also defines a weighted version of $F$, placing
different degrees of importance on precision and recall. This
carries over immediately to yield weighted versions of both
$F'$ and $F^*$.


\section{Discussion}

The overriding concern when choosing a measure of performance in
supervised classification problems should be to match the measure to
the objective. Different measures have different properties,
emphasising different aspects of classification algorithm
performance. A poor choice of measure can lead to the adoption of
an inappropriate classification algorithm, in turn leading to
suboptimal decisions and actions.

A distinguishing characteristic of the F-measure is that it makes
no use of the $TN$ count in the confusion matrix -- the number of class
0 objects correctly classified as class 0. This can be appropriate
in certain domains, such as information retrieval
where $TN$ corresponds to irrelevant documents that are not
retrieved, fraud detection where the number of unflagged legitimate
transactions might be huge, data linkage where there generally is a
large number of correctly unmatched record pairs that are not of
interest, and numerical taxonomy where there is an unlimited number
of characteristics which do not match for any pair of objects. In
other contexts, however, such as in medical diagnosis, correct
classification to each of the classes can be important. 

The F-measure uses the harmonic mean to combine precision and
recall, two distinct aspects of classification algorithm
performance, and some researchers question the use of this form of
mean and the interpretability of their combination. In this paper,
we have shown that suitable transformations of $F$ have
straightforward and familiar intuitive interpretations. Other work
exploring the combination of precision and recall
includes~\citet{Gou05}, \citet{Pow11}, \citet{Boy13},
and~\citet{Fla15}.

\section*{Acknowledgements}

We are grateful to Peter Flach and the two anonymous reviewers
for their helpful comments on an earlier version of this paper.



\bibliography{paper}

\end{document}